\documentclass{article}

    \usepackage[nonatbib,preprint]{neurips_2021}

\usepackage[utf8]{inputenc} 
\usepackage[T1]{fontenc}    
\usepackage{hyperref}       
\usepackage{url}            
\usepackage{booktabs}       
\usepackage{amsfonts}       
\usepackage{nicefrac}       
\usepackage{microtype}      
\usepackage{xcolor}         

\usepackage{caption}
\usepackage{graphicx}
\usepackage{enumitem} 
\usepackage{xspace}
\usepackage{amsmath}
\usepackage{amssymb}
\usepackage{multirow}
\usepackage{algorithm}
\usepackage{algorithmic}
\usepackage{wrapfig}
\usepackage{subcaption}
\usepackage{bm}
\usepackage{listings}

\usepackage{lipsum}

\usepackage{mathtools, nccmath}

\newcommand{\diag}{\operatorname{diag}}

\usepackage{color}

\title{Self-supervised Semi-supervised Learning for Data Labeling and Quality Evaluation}

\author{%
    Haoping Bai\thanks{Corresponding author: Haoping Bai \texttt{haoping\_bai@apple.com}.} \quad Meng Cao \quad Ping Huang \quad Jiulong Shan\\[0.2cm]
    Apple\\[0.2cm]
    \small{\texttt{\{\href{mailto:haoping_bai@apple.com}{haoping\_bai},
    \href{mailto:mengcao@apple.com}{mengcao},
    \href{mailto:huang_ping@apple.com}{huang\_ping}, \href{mailto:jlshan@apple.com}{jlshan}\}@apple.com}}\\
}

\begin{document}

\maketitle

\begin{abstract}
   As the adoption of deep learning techniques in industrial applications grows with increasing speed and scale, successful deployment of deep learning models often hinges on the availability, volume, and quality of annotated data. In this paper, we tackle the problems of efficient data labeling and annotation verification under the human-in-the-loop setting. We showcase that the latest advancements in the field of self-supervised visual representation learning can lead to tools and methods that benefit the curation and engineering of natural image datasets, reducing annotation cost and increasing annotation quality. We propose a unifying framework by leveraging self-supervised semi-supervised learning and use it to construct workflows for data labeling and annotation verification tasks. We demonstrate the effectiveness of our workflows over existing methodologies. On active learning task, our method achieves $97.0\%$ Top-1 Accuracy on CIFAR10 with $0.1\%$ annotated data, and $83.9\%$ Top-1 Accuracy on CIFAR100 with $10\%$ annotated data. When learning with $50\%$ of wrong labels, our method achieves $97.4\%$ Top-1 Accuracy on CIFAR10 and $85.5\%$ Top-1 Accuracy on CIFAR100.

\end{abstract}
\section{Introduction} \label{sec:intro}
As deep learning models and algorithms continue to evolve with increasing capacity and complexity, existing research has shown success across a plethora of domains and tasks~\cite{dosovitskiy2021image, radford2021learning, brock2019large} at the cost of a growing amount of data and compute. However, many industrial applications do not have readily available high-quality datasets. As a result, a large part of the machine learning life cycle is data engineering~\cite{10.1145/3035918.3054782}, which often requires painstaking manual annotation and inspection that are expensive and time-consuming~\cite{blank2020automatic}.

To reduce the amount of human effort, it is necessary to automate the data curation process and reduce the number of labels needed for good performance. For example, active learning can reduce the amount of manual labor required by prioritizing the most informative data sample for labeling. Recent progress in active learning has shown promising results in speeding up human-in-the-loop annotation~\cite{gu2014modelchange,yang2015multi,guo2010nips,dutt2016active,sener2018active}. To help model learn with fewer labels, both self-supervised learning~\cite{chen2020SimCLR,he2020MOCO,chen2020MOCOv2,NEURIPS2020BYOL,caron2020SwAV,chen2021SimSiam} and semi-supervised learning~\cite{NIPS2015_378a063b, NIPS2014_d523773c, odena2016semisupervised, DBLP:conf/iclr/LaineA17, miyato2018virtual, berthelot2019mixmatch, ijcai2019-504, athiwaratkun2018improving, liu2018deep, NIPS2017_68053af2, pmlr-v119-henaff20a,liao2021good} has shown substantial progress, achieving competitive results against supervised baselines with limited supervisions.

Inspired by the latest advances in self-supervised learning and semi-supervised learning, we combine the best of both worlds and build a simple yet versatile image similarity-based framework for robust and efficient labeling and data verification. The contributions of the paper are

\begin{itemize}[leftmargin=10pt]
    
    \item We leverage the latest advances in self-supervised learning and computer vision architectures to obtain image representations that is useful for versatile downstream usages.
    
    \item We demonstrate the effectiveness of the self-supervised representation in a variety of scenarios including active learning-based human-in-the-loop annotation, label error detection, and robust classification with noisy labels. 
    
    \item Unlike expensive semi-supervised approaches that require updating neural networks to incorporate label information~\cite{liao2021good}, our approach leverages label propagation based on nearest neighbor graph to quickly incorporate new label information via simple matrix multiplication. As a result, our method can be seamlessly incorporated into real-time human-in-the-loop systems without sacrificing throughput.
\end{itemize}

We hope our approach will be a simple and strong baseline to motivate future progress in building labor-efficient data curation pipelines and label-efficient machine learning systems. Code will be made available.
\section{Methods}
Our general workflow consists of two parts. We first leverage self-supervised learning, specifically, contrastive learning methods to obtain an unsupervised representation for the unlabeled data. Then we construct a nearest neighbor graph over data samples based on the learned representations. Finally, we can use the nearest neighbor graph for various downstream tasks. In this section, we will discuss each component of our method in detail.

\subsection{Problem formulation}
We assume a dataset of $n$ examples $X = \{\bm{x}_1, \ldots, \bm{x}_n\} \in \mathbb{R}^{n \times d}$ with $d$ being the feature dimension. For our purpose, we define $l \subseteq [n]$ to be the index set for samples with human verified labels and $u \in [n]$ to be the index set for data samples without trusted labels. The label matrix $Y = \{\bm{y}_1, \ldots, \bm{y}_n\} \in \mathbb{R}^{n \times c}$ with $c$ being the number of classes. Our goal is to leverage feature matrix $X$ and known label matrix $Y_l$ to generate and improve the estimate $\tilde{Y}_u$ for the unknown label matrix $Y_u$.

\subsection{Self-supervised Learning}
Recent developments in self-training have seen a substantial progress exemplified by a series of work~\cite{chen2020SimCLR, NEURIPS2020BYOL, he2020MOCO, caron2020SwAV, chen2021SimSiam} in contrastive learning, where the goal is to learn representation that is invariant across two views of the same image created via data augmentation. Specifically, we leverage BYOL~\cite{NEURIPS2020BYOL}, which uses an asymetric siamese architecture including online encoder $f_{\theta}$, online projector $g_{\theta}$ and predictor $q_{\theta}$ for one branch and a target encoder $f_{\xi}$ and projector $g_{\xi}$ for the other branch with polyak averaged weights $\xi \leftarrow \tau \xi + (1-\tau) \theta$. Given two views $v_1$ and $v_2$ of the same image $x$, we obtain projections $p_i = g_{\xi} \circ f_{\xi}(v_i)$ and predictions $z_i = q_{\theta} \circ g_{\theta} \circ f_{\theta}(v_i)$ for $i \in \{1,2\}$, and we train $\theta$ with the following loss

\vspace{-10pt}
\newcommand{\lnorm}[1]{\frac{#1}{\left\lVert{#1}\right\rVert _2}}
\newcommand{\lnormv}[1]{{#1}/{\left\lVert{#1}\right\rVert _2}}
\begin{equation}
    \mathcal{L} = \ell(p_1, z_2)/2 + \ell(p_2, z_1)/2, \quad \text{where} \,\, \ell(p_1, z_2) = - \lnorm{p_1}{\cdot}\lnorm{z_2}
\label{eq:byol}
\end{equation}
\vspace{-10pt}

After training is complete, we use $\ell(f_{\theta}(x_i), f_{\theta}(x_j))$ as a similarity metric between $x_i$ and $x_j$.

\subsection{Nearest Neighbor Graph}
Based on the metric $\ell(f_{\theta}(x_i), f_{\theta}(x_j))$, we can build a nearest neighbor graph in the form of sparse adjacency matrix $W$ where 

\vspace{-10pt}
\begin{align}\label{eq:adj_mat}
	W_{ij} =
	\left\{
		\begin{array}{ll}
			\exp(\ell(f_{\theta}(x_i), f_{\theta}(x_j)) / T), & \mbox{if} \,\, j \in \textrm{NN}(i, k) \\
			0, & \textrm{otherwise}
	   \end{array}
	\right.
\end{align}
\vspace{-10pt}

where $\textrm{NN}(i, k)$ denotes the index set of the k nearest neighbors of $x_i$, and $T$ is a temperature parameter. The symmetrically normalized counterpart of $W$ is given by $\mathcal{W} = D^{-1/2} W D^{-1/2}$, where $D = \diag (W\mathbf{1}_n)$ is the degree matrix and $\mathbf{1}_n$ is an all-ones $n$-vector.

\subsection{Semi-supervised Pseudo-Labeling with Label Propagation}
Based on the consistency assumption~\cite{DBLP:conf/nips/ZhouBLWS03} that nearby nodes are likely to have the same label, we can perform label propagation (LP) on the nearest neighbor graph to propagate information from samples with known labels to samples without label or with noisy labels as follows

\vspace{-10pt}
\begin{equation}
    \tilde{Y}^{(t+1)} = \mathcal{W} Y^{(t)}, \,\, Y^{(t+1)}_{u} = \tilde{Y}^{(t+1)}_{u}, \, Y^{(t+1)}_{l} = Y^{(0)}_{l}
\label{eq:lp}
\end{equation}
\vspace{-15pt}

where $Y^{(0)} = \{ \bm{y}_1, \cdots, \bm{y}_n \}$ is the initial label matrix, $l \subseteq [n]$ denotes index set for samples with annotated/verified labels, $u \subseteq [n]$ denotes index set for samples without trusted labels. $\tilde{Y}^{(t+1)}$ is the soft label matrix at iteration $t+1$, from which we take $\tilde{Y}^{(t+1)}_{u}$ as new soft label for the unlabeled split $u$ and reset $ Y^{(t+1)}_{l}$ to the given ground truth $Y^{(0)}_{l}$. Since LP is mostly sparse matrix multiplication, incorporating information from added labels into the pseudo-labels is much faster than training a deep learning model with partial labels.

\section{Experimental Analysis and Results} \label{sec:exp}
In this section, we showcase the performance of two convenient workflows that we built around the aforementioned techniques with minimal modifications.

\noindent \textbf{Experiment Settings and Implementation Details} We use CIFAR10 and CIFAR100~\cite{Krizhevsky2009LearningML} as our benchmark datasets. To perform self-supervised learning on target dataset, we leverage the Vision Transformer~\cite{dosovitskiy2021image}, ViT-B/16, as our encoder architecture while following the exact projector and predictor definition in~\cite{dosovitskiy2021image}. We initialize the ViT encoder with BEiT~\cite{beit} pretrained weights. We perform all training with batchsize 64 with images resized to $224 \times 224$ resolution. We use the AdamW optimizer~\cite{loshchilov2018decoupled}. For each experiment, we determine the learning rate, weight decay, and $\tau$ through grid search. On CIFAR10, we run BYOL for 3k steps with a learning rate of $2.3e-5$, $5.e-4$ weight decay, $\tau = 0.998$. On CIFAR100, we run BYOL for 5k steps with a learning rate of $4.6e-5$, $2.1e-6$ weight decay, $\tau = 0.9993$.  Following~\cite{dosovitskiy2021image}, we gradually anneal $\tau$ to 1 during training. To construct k-NN graph, we use $k=10$ and $T=0.01$ for CIFAR10, and $k=15$ and $T=0.02$ for CIFAR100. We use $t=20$ iterations for LP.

\noindent \textbf{k-NN Classification Performance with Learned Representations} To assess the quality of our learned representation, we directly performed weighted k-NN classification based on the nearest neighbor graph in Equation~\ref{eq:adj_mat}. We achieved $98.45\%$ validation Top-1 accuracy on CIFAR10, and $89.58\%$ validation Top-1 accuracy on CIFAR100. 


\vspace{-5pt}
\begin{figure}[ht!]
    \centering
    \small
    \includegraphics[width=0.98\linewidth]{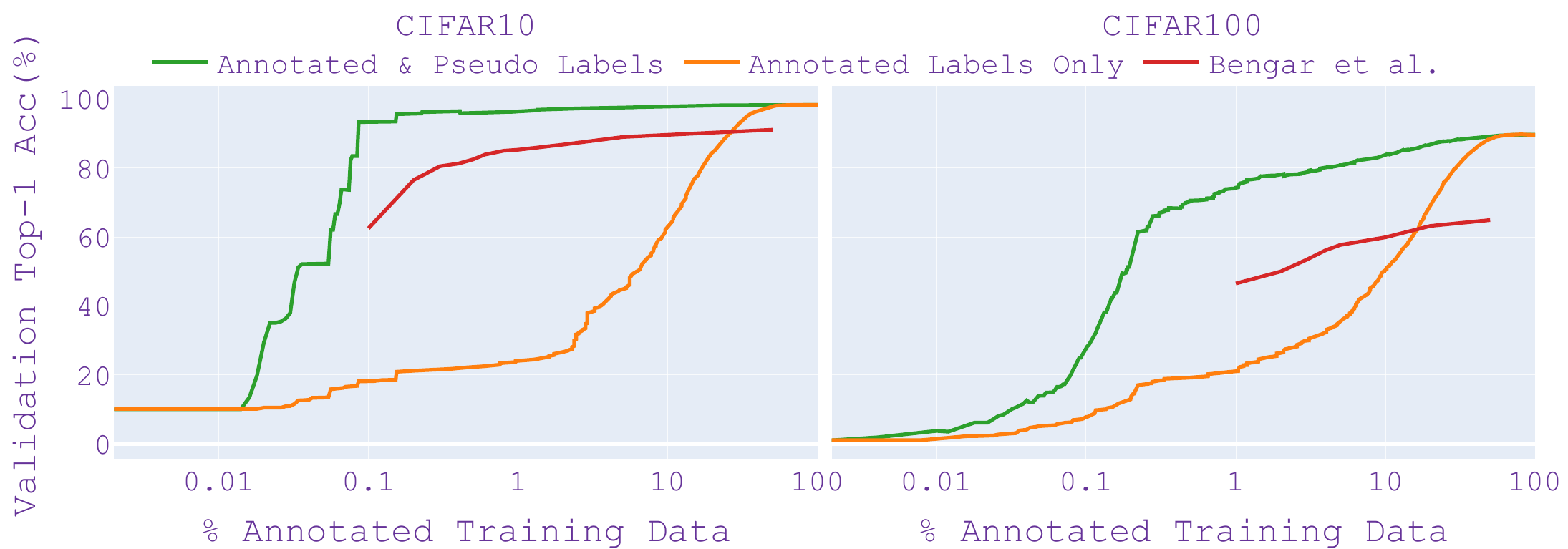}
    \vspace{-4pt}
    \caption{Active Learning Performance on CIFAR10 (left) and CIFAR100 (right). Orange line denotes using only labeled training data to predict validation labels. Red line denotes results from Bengar et al.~\cite{bengar2021reducing}. Green line denotes using both labeled training data and the unlabeled training data with LP generated pseudo-labels to predict validation labels.
    }
    \label{fig:n_ann_v_acc}
    \vspace{-9pt}
\end{figure}

\subsection{Efficient Annotation with Active Learning}
For this task, we perform simulation using both datasets and start with no training label. We simulate the human-in-the-loop annotation process by iteratively performing LP and randomly sampling data for oracle labeling. Following the observation in~\cite{bengar2021reducing}, we choose the random sampling as a simple and strong baseline. As shown in Figure~\ref{fig:n_ann_v_acc}, we achieve exponential gain in Top-1 Accuracy when annotating fewer than $< 0.1\%$ data in CIFAR10 and $< 1\%$ data in CIFAR100. We outperform previous work by Bengar et al.~\cite{bengar2021reducing} substantially.

To assess the value of pseudo-labels generated by LP on the unlabeled training data, we perform an ablation study by performing LP only on the annotated data and the validation data. As shown in Figure~\ref{fig:n_ann_v_acc}, without pseudo-labels from the unlabeled training data, the orange curves show substantial drops in validation top-1 accuracy when using the same amount of annotation. Thus, having a reliable nearest neighbor graph allows us to effectively scale performance with the amount of unlabeled data by propagating information from labeled data across the data manifold.

\begin{figure} 
    \centering
    \small
    \includegraphics[width=0.98\linewidth]{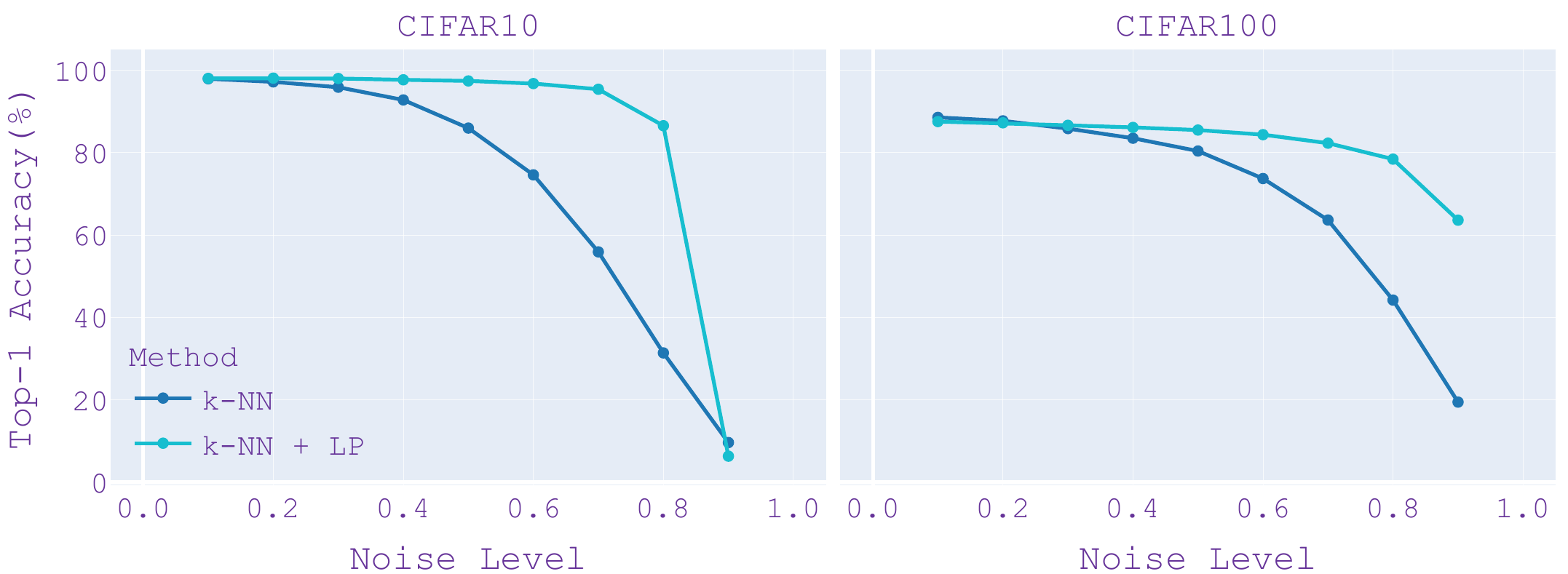}
    \vspace{-4pt}
    \caption{Classification Performance Under Label Noise on CIFAR10 (left) and CIFAR100 (right)}
    \label{fig:noise_v_acc}
    \vspace{-9pt}
\end{figure}

\subsection{Robust Classification with Noisy Labels}

We showcase our second workflow by demonstrating the effectiveness of using LP to correct corrupted labels and maintain robust classification performance under noise.

\begin{wrapfigure}{r}{0.45\textwidth}
	\centering
 	\vspace{-15pt}
    \includegraphics[width=\linewidth]{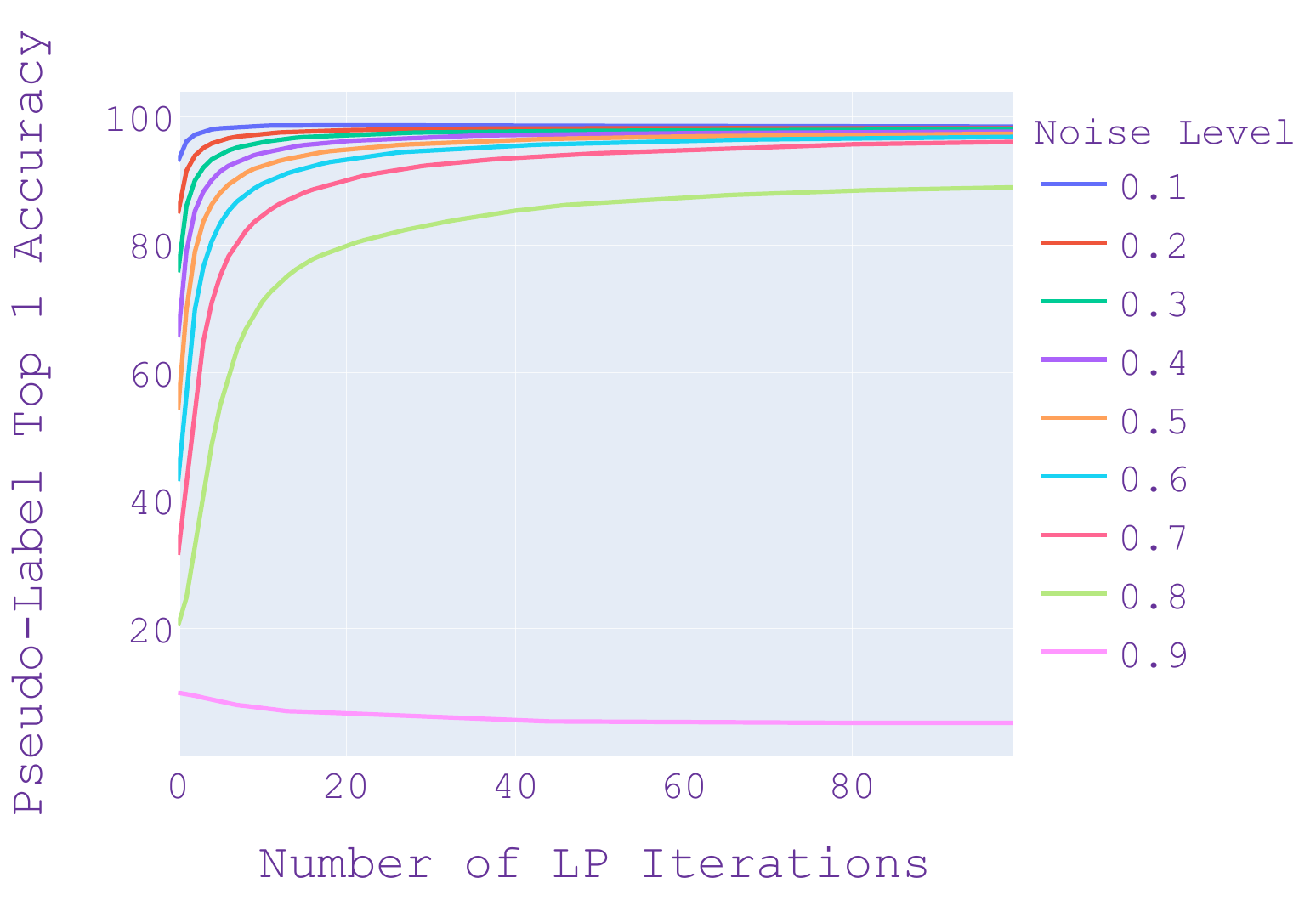}
    \captionof{figure}{The Noise Level in Pseudo Label vs. Number of LP Iterations}
    \label{fig:lp_vs_noise}
    \vspace{-15pt}
\end{wrapfigure}

\noindent \textbf{Effect of LP on Noise Reduction} We first examine the evidence for the consistency assumption~\cite{DBLP:conf/nips/ZhouBLWS03} by simulating random noise in labels and then performing LP. If our nearest neighbor graph faithfully captures the similarity among data, LP will aggregate and smooth out the inconsistency from noisy neighbor labels so that the neighbors with the correct label can stand out. As illustrated in Figure~\ref{fig:lp_vs_noise}, with a noise level below 0.8, LP quickly reduces the noise level in pseudo label with an increasing number of iterations. At extreme noise levels such as 0.9, the consistency assumption no longer holds due to highly corrupted labels, and performing LP hurts performance instead.

\noindent \textbf{Classification under Noisy Label} After obtaining smoothed pseudo labels via LP, we use the pseudo labels with the nearest neighbor graph to perform weighted k-NN classification. For comparison, we use weighted k-NN based on the corrupted labels as a baseline. As shown in Figure~\ref{fig:noise_v_acc}, pseudo labels obtained via LP offer more robust performance than directly using the corrupted labels on both CIFAR10 and CIFAR100.

\section{Conclusion}
In summary, leveraging the latest advances in self-supervised learning, we developed a nearest neighbor graph-based approach that can perform versatile downstream tasks and quickly incorporate new information in a semi-supervised manner, suitable for integration with real time human-in-the-loop systems. We demonstrated the effectiveness of our methods on a combination of datasets (CIFAR10, CIFAR100) and tasks (active learning, label error detection, and learning under noise) and achieved competitive performance. We hope our work can serve as a simple and strong baseline for the development of practical tools in industrial settings.
\newpage
\bibliographystyle{unsrt}{\small
\bibliography{egbib}

\begin{thebibliography}{10}

\bibitem{dosovitskiy2021image}
Alexey Dosovitskiy, Lucas Beyer, Alexander Kolesnikov, Dirk Weissenborn,
  Xiaohua Zhai, Thomas Unterthiner, Mostafa Dehghani, Matthias Minderer, Georg
  Heigold, Sylvain Gelly, Jakob Uszkoreit, and Neil Houlsby.
\newblock An image is worth 16x16 words: Transformers for image recognition at
  scale, 2021.

\bibitem{radford2021learning}
Alec Radford, Jong~Wook Kim, Chris Hallacy, Aditya Ramesh, Gabriel Goh,
  Sandhini Agarwal, Girish Sastry, Amanda Askell, Pamela Mishkin, Jack Clark,
  Gretchen Krueger, and Ilya Sutskever.
\newblock Learning transferable visual models from natural language
  supervision, 2021.

\bibitem{brock2019large}
Andrew Brock, Jeff Donahue, and Karen Simonyan.
\newblock Large scale gan training for high fidelity natural image synthesis,
  2019.

\bibitem{10.1145/3035918.3054782}
Neoklis Polyzotis, Sudip Roy, Steven~Euijong Whang, and Martin Zinkevich.
\newblock Data management challenges in production machine learning.
\newblock In {\em Proceedings of the 2017 ACM International Conference on
  Management of Data}, SIGMOD '17, page 1723–1726, New York, NY, USA, 2017.
  Association for Computing Machinery.

\bibitem{blank2020automatic}
Clas Blank.
\newblock Automatic vs. manual data labeling: A system dynamics modeling
  approach, 2020.

\bibitem{gu2014modelchange}
Wenbin Cai, Ya~Zhang, Siyuan Zhou, Wenquan Wang, Chris Ding, and Xiao Gu.
\newblock Active learning for support vector machines with maximum model
  change.
\newblock In {\em Machine Learning and Knowledge Discovery in Databases}, pages
  211--226. Springer, 2014.

\bibitem{yang2015multi}
Yi~Yang, Zhigang Ma, Feiping Nie, Xiaojun Chang, and Alexander~G Hauptmann.
\newblock Multi-class active learning by uncertainty sampling with diversity
  maximization.
\newblock {\em IJCV}, 113(2):113--127, 2015.

\bibitem{guo2010nips}
Yuhong Guo.
\newblock Active instance sampling via matrix partition.
\newblock In {\em NIPS}, pages 1--9, 2010.

\bibitem{dutt2016active}
Suyog Dutt~Jain and Kristen Grauman.
\newblock Active image segmentation propagation.
\newblock In {\em CVPR}, pages 2864--2873, 2016.

\bibitem{sener2018active}
Ozan Sener and Silvio Savarese.
\newblock Active learning for convolutional neural networks: A core-set
  approach.
\newblock In {\em ICLR}, 2018.

\bibitem{chen2020SimCLR}
Ting Chen, Simon Kornblith, Mohammad Norouzi, and Geoffrey Hinton.
\newblock A simple framework for contrastive learning of visual
  representations.
\newblock In {\em International conference on machine learning}, pages
  1597--1607. PMLR, 2020.

\bibitem{he2020MOCO}
Kaiming He, Haoqi Fan, Yuxin Wu, Saining Xie, and Ross Girshick.
\newblock Momentum contrast for unsupervised visual representation learning.
\newblock In {\em Proceedings of the IEEE/CVF Conference on Computer Vision and
  Pattern Recognition}, pages 9729--9738, 2020.

\bibitem{chen2020MOCOv2}
Xinlei Chen, Haoqi Fan, Ross Girshick, and Kaiming He.
\newblock Improved baselines with momentum contrastive learning.
\newblock {\em arXiv preprint arXiv:2003.04297}, 2020.

\bibitem{NEURIPS2020BYOL}
Jean-Bastien Grill, Florian Strub, Florent Altch\'{e}, Corentin Tallec, Pierre
  Richemond, Elena Buchatskaya, Carl Doersch, Bernardo Avila~Pires, Zhaohan
  Guo, Mohammad Gheshlaghi~Azar, Bilal Piot, koray kavukcuoglu, Remi Munos, and
  Michal Valko.
\newblock Bootstrap your own latent - a new approach to self-supervised
  learning.
\newblock In H.~Larochelle, M.~Ranzato, R.~Hadsell, M.~F. Balcan, and H.~Lin,
  editors, {\em Advances in Neural Information Processing Systems}, volume~33,
  pages 21271--21284. Curran Associates, Inc., 2020.

\bibitem{caron2020SwAV}
Mathilde Caron, Ishan Misra, Julien Mairal, Priya Goyal, Piotr Bojanowski, and
  Armand Joulin.
\newblock Unsupervised learning of visual features by contrasting cluster
  assignments.
\newblock {\em arXiv preprint arXiv:2006.09882}, 2020.

\bibitem{chen2021SimSiam}
Xinlei Chen and Kaiming He.
\newblock Exploring simple siamese representation learning.
\newblock In {\em Proceedings of the IEEE/CVF Conference on Computer Vision and
  Pattern Recognition}, pages 15750--15758, 2021.

\bibitem{NIPS2015_378a063b}
Antti Rasmus, Mathias Berglund, Mikko Honkala, Harri Valpola, and Tapani Raiko.
\newblock Semi-supervised learning with ladder networks.
\newblock In C.~Cortes, N.~Lawrence, D.~Lee, M.~Sugiyama, and R.~Garnett,
  editors, {\em Advances in Neural Information Processing Systems}, volume~28.
  Curran Associates, Inc., 2015.

\bibitem{NIPS2014_d523773c}
Durk~P Kingma, Shakir Mohamed, Danilo Jimenez~Rezende, and Max Welling.
\newblock Semi-supervised learning with deep generative models.
\newblock In Z.~Ghahramani, M.~Welling, C.~Cortes, N.~Lawrence, and K.~Q.
  Weinberger, editors, {\em Advances in Neural Information Processing Systems},
  volume~27. Curran Associates, Inc., 2014.

\bibitem{odena2016semisupervised}
Augustus Odena.
\newblock Semi-supervised learning with generative adversarial networks, 2016.

\bibitem{DBLP:conf/iclr/LaineA17}
Samuli Laine and Timo Aila.
\newblock Temporal ensembling for semi-supervised learning.
\newblock In {\em 5th International Conference on Learning Representations,
  {ICLR} 2017, Toulon, France, April 24-26, 2017, Conference Track
  Proceedings}. OpenReview.net, 2017.

\bibitem{miyato2018virtual}
Takeru Miyato, Shin ichi Maeda, Masanori Koyama, and Shin Ishii.
\newblock Virtual adversarial training: A regularization method for supervised
  and semi-supervised learning, 2018.

\bibitem{berthelot2019mixmatch}
David Berthelot, Nicholas Carlini, Ian Goodfellow, Nicolas Papernot, Avital
  Oliver, and Colin Raffel.
\newblock Mixmatch: A holistic approach to semi-supervised learning, 2019.

\bibitem{ijcai2019-504}
Vikas Verma, Alex Lamb, Juho Kannala, Yoshua Bengio, and David Lopez-Paz.
\newblock Interpolation consistency training for semi-supervised learning.
\newblock In {\em Proceedings of the Twenty-Eighth International Joint
  Conference on Artificial Intelligence, {IJCAI-19}}, pages 3635--3641.
  International Joint Conferences on Artificial Intelligence Organization, 7
  2019.

\bibitem{athiwaratkun2018improving}
Ben Athiwaratkun, Marc Finzi, Pavel Izmailov, and Andrew~Gordon Wilson.
\newblock There are many consistent explanations of unlabeled data: Why you
  should average.
\newblock {\em ICLR}, 2019.

\bibitem{liu2018deep}
Bin Liu, Zhirong Wu, Han Hu, and Stephen Lin.
\newblock Deep metric transfer for label propagation with limited annotated
  data.
\newblock 2018.

\bibitem{NIPS2017_68053af2}
Antti Tarvainen and Harri Valpola.
\newblock Mean teachers are better role models: Weight-averaged consistency
  targets improve semi-supervised deep learning results.
\newblock In I.~Guyon, U.~V. Luxburg, S.~Bengio, H.~Wallach, R.~Fergus,
  S.~Vishwanathan, and R.~Garnett, editors, {\em Advances in Neural Information
  Processing Systems}, volume~30. Curran Associates, Inc., 2017.

\bibitem{pmlr-v119-henaff20a}
Olivier Henaff.
\newblock Data-efficient image recognition with contrastive predictive coding.
\newblock In Hal~Daumé III and Aarti Singh, editors, {\em Proceedings of the
  37th International Conference on Machine Learning}, volume 119 of {\em
  Proceedings of Machine Learning Research}, pages 4182--4192. PMLR, 13--18 Jul
  2020.

\bibitem{liao2021good}
Yuan-Hong Liao, Amlan Kar, and Sanja Fidler.
\newblock Towards good practices for efficiently annotating large-scale image
  classification datasets, 2021.

\bibitem{DBLP:conf/nips/ZhouBLWS03}
Dengyong Zhou, Olivier Bousquet, Thomas~Navin Lal, Jason Weston, and Bernhard
  Schölkopf.
\newblock Learning with local and global consistency.
\newblock In {\em NIPS}, pages 321--328, 2003.

\bibitem{Krizhevsky2009LearningML}
A.~Krizhevsky.
\newblock Learning multiple layers of features from tiny images.
\newblock 2009.

\bibitem{beit}
Hangbo Bao, Li~Dong, and Furu Wei.
\newblock {BEiT}: {BERT} pre-training of image transformers.
\newblock 2021.

\bibitem{loshchilov2018decoupled}
Ilya Loshchilov and Frank Hutter.
\newblock Decoupled weight decay regularization.
\newblock In {\em International Conference on Learning Representations}, 2019.

\bibitem{bengar2021reducing}
Javad~Zolfaghari Bengar, Joost van~de Weijer, Bartlomiej Twardowski, and Bogdan
  Raducanu.
\newblock Reducing label effort: Self-supervised meets active learning, 2021.

\end{thebibliography}
}

\end{document}